\pdfoutput=1

\documentclass[11pt]{article}

\usepackage[preprint]{acl}
\usepackage{graphicx}
\usepackage{booktabs}  
\usepackage{multirow}  
\usepackage{adjustbox} 

\usepackage{times}
\usepackage{tcolorbox}
\usepackage{array}
\usepackage{tabularx}
\usepackage{fancybox}
\usepackage{latexsym}
\usepackage{amsmath}
\usepackage{amssymb}
\usepackage{verbatim}
\usepackage[T1]{fontenc}

\usepackage[utf8]{inputenc}

\usepackage{microtype}

\usepackage{inconsolata}

\usepackage{graphicx}

\newcommand{\nan}[1]{\textcolor{blue}{{nan: #1}}}
\hypersetup{draft}

\title{Review-Then-Refine: A Dynamic Framework for Multi-Hop Question Answering with Temporal Adaptability}

\author{Xiangsen Chen \ Xuming Hu\footnotemark[1] \  Nan Tang\footnotemark[1] \\
        Hong Kong University of Science and Technology(Guangzhou) \\
        xchen748@connect.hkust-gz.edu.cn\\ xuminghu97@gmail.com\\ nantang@hkust-gz.edu.cn\\
        }

\begin{document}
\maketitle
\begin{abstract}
Retrieve-augmented generation (RAG) frameworks have emerged as a promising solution to multi-hop question answering(QA) tasks since it enables large language models (LLMs) to incorporate external knowledge and mitigate their inherent knowledge deficiencies. Despite this progress, existing RAG frameworks, which usually follows the retrieve-then-read paradigm, often struggle with multi-hop QA with temporal information since it has difficulty retrieving and synthesizing accurate time-related information. 
To address the challenge, this paper proposes a novel framework called \textit{review-then-refine}, which aims to enhance LLM performance in multi-hop QA scenarios with temporal information. 
Our approach begins with a review phase, where decomposed sub-queries are dynamically rewritten with temporal information
, allowing for subsequent adaptive retrieval and reasoning process. In addition, we implement adaptive retrieval mechanism to minimize unnecessary retrievals, thus reducing the potential for hallucinations. In the subsequent refine phase, the LLM synthesizes the retrieved information from each sub-query along with its internal knowledge to formulate a coherent answer. Extensive experimental results across multiple datasets demonstrate the effectiveness of our proposed framework, highlighting its potential to significantly improve multi-hop QA capabilities in LLMs.

\end{abstract}
\footnotetext{Correspondence to Xuming Hu: xuminghu97@gmail.com and Nan Tang: nantang@hkust-gz.edu.cn}

\section{Introduction}

\begin{figure*}[htbp]
    \centering
    \includegraphics[width=\textwidth]{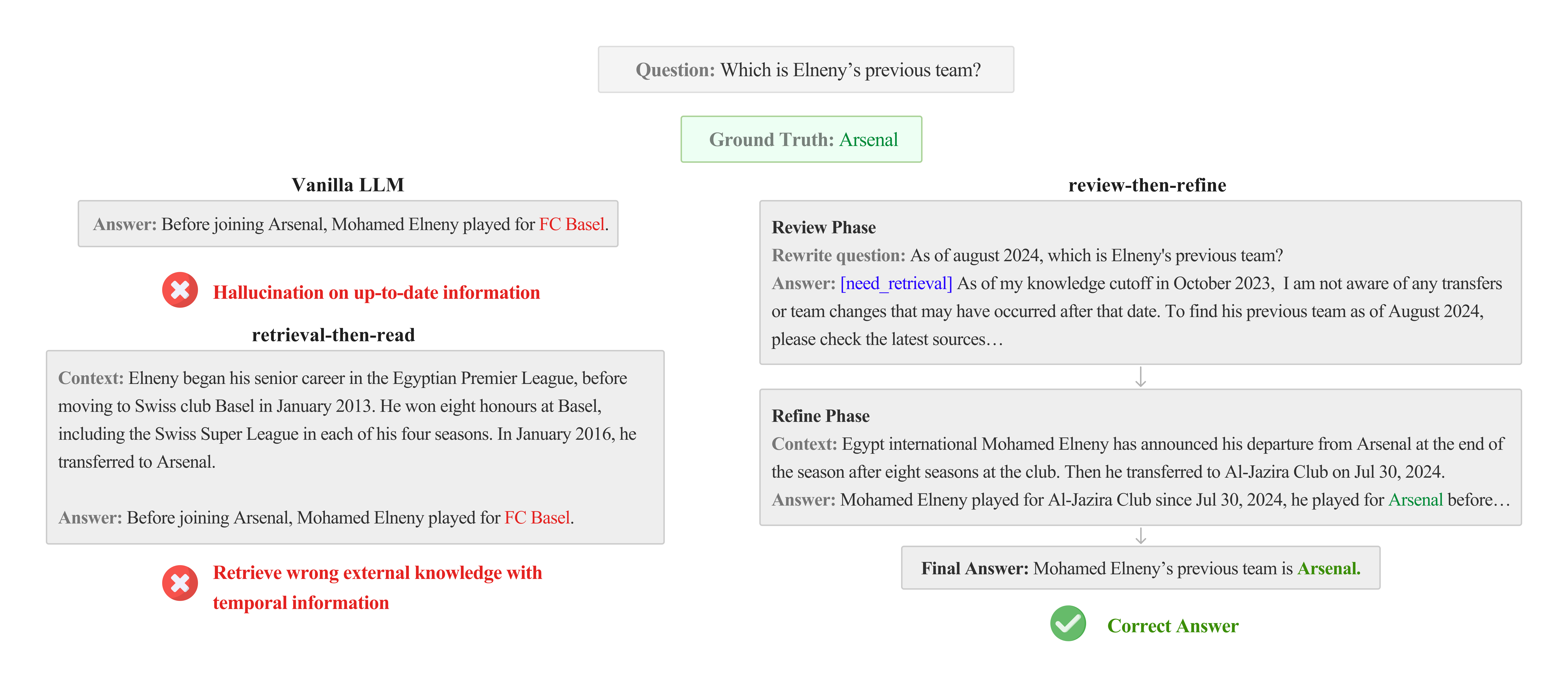}
    \caption{Illustration of challenges in multi-hop QA with temporal information and how our proposed \textit{review-then-refine} method addresses them. The traditional retrieval-then-read paradigm often introduces external hallucinations or retrieves wrong information, while the review-then-refine method dynamically adjusts retrieval and query rewriting to produce the correct answer.}
    \label{fig:challenge_solution} 
\end{figure*}

Multi-hop question answering(QA) has been extensively applied in various practical domains \cite{mavi2022survey, chen2017readingopen-domain, feng2022multi} and it requires models to retrieve and integrate information from multiple sources, and retrieve-augmented generation (RAG) is a crucial method in enhancing large language models (LLMs) for multi-hop question answering. \cite{ragqasiriwardhana2023improving} 
Previous works have achieved significant progress in multi-hop question answering with their internal knowledge to generate coherent and contextually relevant text. \cite{mavi2022survey, ragqasiriwardhana2023improving, Verify-and-editzhao2023verify} And the methods have predominantly adopted the retrieve-then-read paradigm \cite{shao-etal-2023-enhancingretrive-then-read}. When given a query, this approach generally involves two key steps: \textbf{Retrieve:} Extracting all relevant contexts from an external corpus, and \textbf{Read:} Analyzing the retrieved contexts to extract or infer the correct answer. 



While this framework has shown considerable success in multi-hop QA with iterative process, \cite{wei2022cot, decomposepatel2022question, multireasoningchainschen2019multi} however, they still face challenges in accurately answering complex, multi-hop questions, especially those with temporal information \cite{vu2023freshllms}. Additional challenges under-explored are shown in Figure \ref{fig:challenge_solution}. 
Each step in the iterative process introduces potential failure points, where the irrelevant or incorrect data may introduced, leading to erroneous conclusions. 
This approach is inadequate for handling time-sensitive queries, where answers must rely on the latest information with respect to the query \cite{cheng2024multi-temporal}. Therefore, the external information may not contain the correct temporal data, and the inclusion of external information increases the risk of external hallucinations \cite{yan2024corrective, ding2024retrievehallucination}, where irrelevant or misleading data contaminates the model’s internal knowledge, resulting in biased or incorrect answers \cite{wu2024gendec}. Even for methods with advanced fine-tuning 
techniques, they still struggle to provide reliable answers to questions that require temporal awareness or up-to-date information \cite{cheng2024multi-temporal, jeong2024adaptiverag, vu2023freshllms}.

To overcome the challenges inherent in the retrieve-then-read paradigm, particularly in the context of multi-hop question answering with temporal information, we propose a novel framework called \textit{review-then-refine}. This framework is specifically designed to enhance the performance of LLMs by addressing the key shortcomings of existing approaches through a two-phase process:
(1) \textbf{Review Phase:} The traditional retrieve-then-read approach often struggles with complex temporal queries, resulting in both retrieval and synthesis errors. In the review phase, the framework breaks down the original query into simpler sub-queries with more accurate temporal data to generate a reasoning chain, thus enabling more adaptive retrieval of relevant information. This decomposition allows for better navigation of multi-hop reasoning. Furthermore, a adaptive retrieval mechanism is implemented to evaluate the necessity of each retrieval step, thereby minimizing the risk of external hallucinations caused by irrelevant data contaminating the model’s internal knowledge base \cite{ding2024retrievehallucination}. (2) \textbf{Refine Phase:} The refine phase addresses the issue of inadequate synthesis in traditional approaches, where the retrieved information may not fully align with the model's internal knowledge. This phase integrates external data with the model's existing understanding, ensuring that the final answer is both accurate and contextually coherent. By verifying external data, the refine phase minimizes hallucinated responses and improves the overall quality and accuracy of the answers generated.

Table \ref{table:comparison} illustrates a comparison between existing methods and our approach, highlighting the unique contributions of \textit{review-then-refine} in key areas such as dynamic query rewriting, query decomposition, and handling temporal queries.
\begin{table*}[htbp]
    \centering
    \caption{Comparison of different multi-hop QA approaches.}
    \label{table:comparison}
    \small
    \begin{adjustbox}{max width=\textwidth}
    \begin{tabular}{>{\centering\arraybackslash}p{2.5cm} >{\centering\arraybackslash}p{2cm} >{\centering\arraybackslash}p{2cm} >{\centering\arraybackslash}p{2cm} >{\centering\arraybackslash}p{2cm} >{\centering\arraybackslash}p{2cm}}
        \toprule
        \textbf{Paper} & \textbf{Query Rewriting} & \textbf{Query Decomposing} & \textbf{Retrieve Adaptability} & \textbf{Contextual Check} & \textbf{Temporal Data} \\
        \midrule
        FLARE\cite{jiang2023flare} & $\times$ & $\times$ & $\checkmark$ & $\times$ & $\times$ \\
        RQ-RAG\cite{chan2024rq-rag} & $\checkmark$ & $\checkmark$ & $\times$ & $\checkmark$ & $\times$ \\
        Self-RAG\cite{selfragasai2023self} & $\times$ & $\checkmark$ & $\checkmark$ & $\checkmark$ & $\times$ \\
        Adaptive-RAG\cite{jeong2024adaptiverag} & $\checkmark$ & $\checkmark$ & $\checkmark$ & $\times$ & $\times$ \\
        \textit{review-then-refine} & $\checkmark$ & $\checkmark$ & $\checkmark$ & $\checkmark$ & $\checkmark$ \\
        \bottomrule
    \end{tabular}
    \end{adjustbox}
\end{table*}
As shown in Table \ref{table:comparison}, our method excels across several dimensions by combining dynamic query rewriting with query decomposition, ensuring contextually relevant retrieval, and handling time-sensitive queries, a capability that many existing approaches lack.

The main contributions of this paper are:
\begin{itemize}
    \item We propose a novel \textit{review-then-refine} framework to address the limitations of existing retrieve-then-read methods in multi-hop QA.
    \item We incorporate mechanisms to handle temporal information and minimize external hallucinations with dynamic rewriting and filtering irrelevant data during retrieval.
    \item The method achieves state-of-the-art performance on several multi-hop QA benchmarks.
\end{itemize}
\section{Related Work}
\subsection{Multi-Hop Question Answering}
Multi-hop question answering (QA) requires models to retrieve and synthesize information across multiple external sources to answer complex queries. \cite{zhang2024end, multireasoningchainschen2019multi}. To tackle the problem, recent work has focused on decomposition strategies \cite{perez2020unsupervised} with the internal knowledge of LLMs \cite{radhakrishnan2023questiondecompose}, which break down complex questions into simpler sub-queries, and answer them step-by-step with external knowledge. \cite{zhou2022Least-to-most, trivedi2022ircot, khot2022decomposed} Besides, there are previous studies that using CoT \cite{wei2022cot} to decompose complex questions into easier sub-questions and answer sub-questions step by step with external context by LLMs. \cite{khattab2022demonstrate, self-askpress2022measuring} 
And XoT-based methods like tree-of-thought(ToT) \cite{yao2024tot} or graph-of-thought(GoT) \cite{besta2024got} are an effective way in resolving more complex questions by generating an reasoning path for each question.
However, despite their progress, CoT methods struggles to generate an accurate answer since it overlooks the potential error in retrieving, and XoT-based methods usually have a higher variance in generating an answer.
In this work, we constructed the review module to generate an effective and accurate reasoning path with dynamic rewriting and decomposing and adaptive retrieval to mitigate this challenge.

\subsection{Retrieval-Augmented Generation for LLMs}
With relevant information retrieved from external knowledge sources, Retrieval-augmented LLMs have been demonstrated as an effective way to improve the performance of LLMs in downstream tasks like multi-hop QA by reducing hallucinations in generated outputs. \cite{khattab2022demonstrate, lazaridou2022internet}  There has been significant research advances in Multi-hop QA with the RAG pipeline, which typically involves the paradigm known as \textit{retrieve-then-read}. \cite{shao-etal-2023-enhancingretrive-then-read, chen2017readingopen-domain}. And previous \textit{retrieve-then-read} approaches tackle Multi-hop QA by combining the context of the retrieved corpus with the internal knowledge of LLMs \cite{abdallah2024generator}, or fine-tune LMs to determine when to retrieve \cite{selfragasai2023self, jeong2024adaptiverag, wang2024selfdc}, with an iterative process.
However, despite the success of \textit{retrieve-then-read} frameworks, the aforementioned methods ignored the fact that questions are in a wide variety of dynamic status \cite{zhao2024set}. And there is temporal conflict of different external knowledge for different temporal information \cite{vu2023freshllms}, thus leading to the potential error in retrieving \cite{xu2024search} and synthesizing multiple pieces of information \cite{yao2022react}, which may hallucinate the LLMs in the reasoning chain. In our work, we propose a novel framework \textit{review-then-refine} that aims to more effectively combine the internal knowledge of LLMs with external context with more accurate temporal information, thus reducing the potential for hallucination and enhancing the accuracy of the answers to various questions with temporal information.



\section{Review-then-Refine}
\begin{figure*}[htbp]
    \centering
    \includegraphics[width=0.95\textwidth]{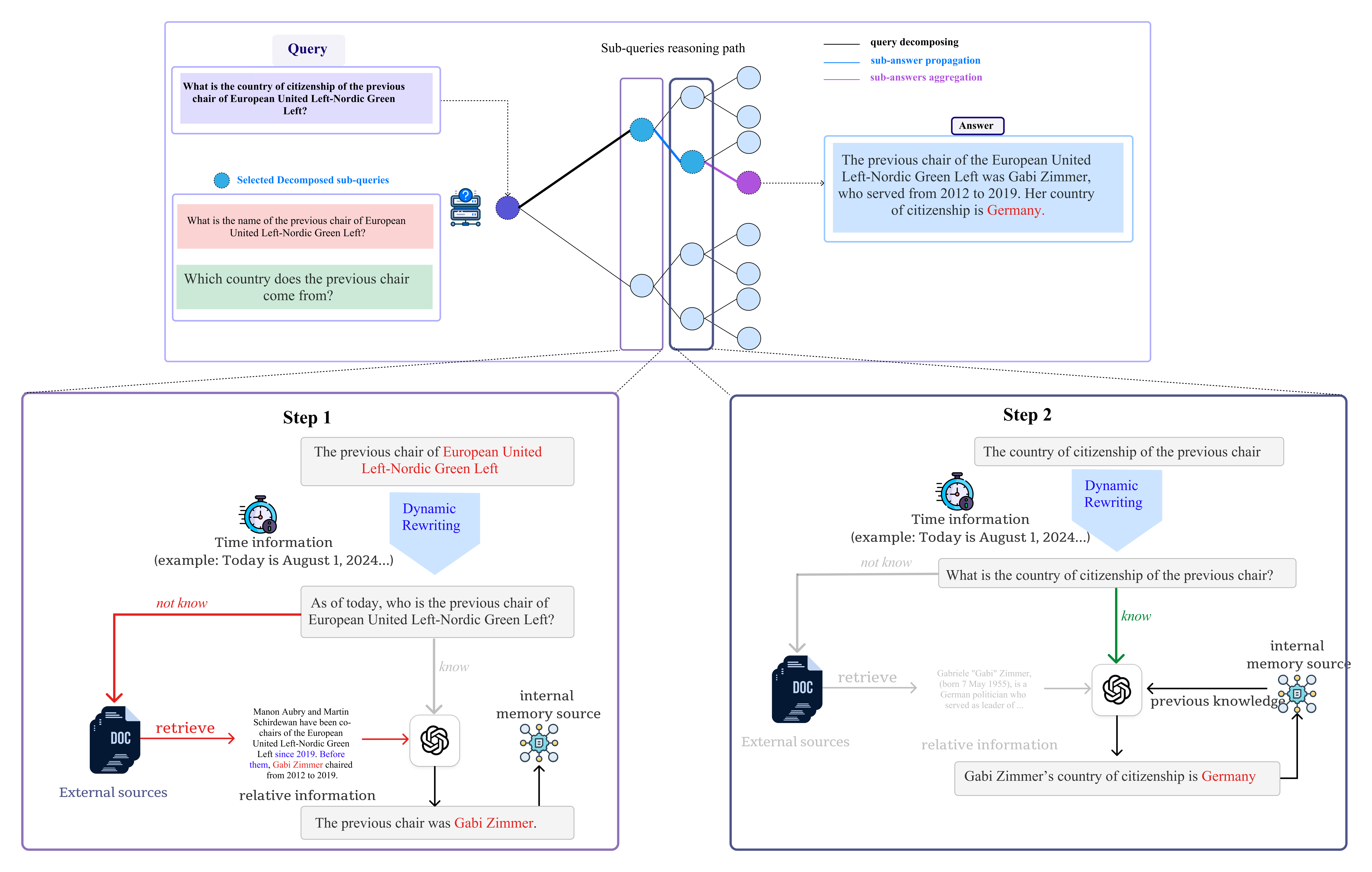}
    \caption{The illustration of how the framework works.}
    \label{Figure 2}  
\end{figure*}

This section provides a detailed explanation of the proposed framework, review-then-refine. This framework comprises two main phases: the review phase and the refine phase. Each phase plays a crucial role in dynamically rewriting queries, integrating temporal data, and synthesizing information from multiple sources into coherent and accurate answers. The illustration of the framework is shown as Figure \ref{Figure 2}.

\subsection{Review}

The review phase is initiated with the dynamic query rewriting component, which is critical for managing multi-hop queries. Unlike traditional static query decomposition methods, this component decomposes a complex multi-hop query \( Q \) into simpler, more manageable sub-queries \( \{q_1, q_2, \ldots, q_n\} \). At its core, query decomposition is a dynamic, iterative process in which the system breaks down the original query \( Q \) based on the current reasoning state. This allows for real-time adjustments to the query as new information is uncovered. The system does not follow a predetermined static path, but rather dynamically rewrites and adapts the query decomposition based on intermediate results. 

Let the original query \( Q \) be represented as \(Q = \{q_0, c_0, t_0\}\), where \( q_0 \) is the query itself, \( c_0 \) is the contextual information available at the start, and \( t_0 \) is the time context (if relevant). Each sub-query \( q_i \) is generated using a dynamic rewriting function \( f_i \) based on the original query \( Q \) and the current reasoning history \( H_i \). The dynamic generation of each sub-query is formalized as \(q_i = f_i(Q, H_i)\) where \( H_i = \{ (q_1, a_1), (q_2, a_2), \dots, (q_{i-1}, a_{i-1}) \} \) represents the history of previous sub-queries and their corresponding answers. Each sub-query \( q_i \) targets a specific aspect of the overall question, breaking it into solvable one-hop queries.

We define the history of sub-queries and their corresponding answers as \( H_i = \{(q_j, \tilde{a}_j) \mid j < i \} \), where \( q_j \) is the sub-query generated at step \( j \), and \( \tilde{a}_j \) is its corresponding answer. In the \( i \)-th iteration, the large language model \( M_{\theta} \) generates the next sub-query \( q_i \) and its anticipated answer \( \tilde{a}_i \) based on the initial query \( Q \) and the accumulated history \( H_i \):
\begin{equation}
q_i, \tilde{a}_i = M_{\theta}(Q, H_i)
\end{equation}

At each step, the model \( M_{\theta} \) dynamically rewrites the query to address the gaps identified in previous steps. For a given sub-query \( q_i \), the system must decide whether to perform external retrieval or rely solely on existing knowledge from previous sub-queries. This decision is processed by evaluating whether the model can generate a strictly accurate answer without hallucination. Specifically, if the model is not fully capable of generating an answer that is completely supported by its internal knowledge, or does not include any incorrect, speculative, or fabricated information, then the system will require external retrieval. This decision is formalized using a retrieval indicator function \( \mathbb{I}(q_i, H_i) \), where \( H_i \) is the reasoning history up to step \( i \):
\[
\mathbb{I}(q_i, H_i) =
\begin{cases}
1, & \text{if \text{external retrieval} is required}, \\
0, & \text{else}
\end{cases}
\]

If \( \mathbb{I}(q_i, H_i) = 1 \), indicating that the model cannot generate a fully accurate answer based solely on its internal knowledge, the system performs a retrieval operation. In this case, a [need\_retrieval] label is added, and the retrieval process is initiated as follows:
\[
D_i = \text{Retrieve}(q_i, \mathbb{I}(q_i, H_i)),
\]
where \( D_i \) represents the set of relevant documents or external sources retrieved to answer \( q_i \). This function \( \text{Retrieve}(q_i, \mathbb{I}(q_i, H_i)) \) invokes a search engine or specialized document retriever when \( \mathbb{I}(q_i, H_i) = 1 \), and returns an empty set if no retrieval is required.

The retrieved documents \( D_i \) are then analyzed for pertinent information to address the current sub-query \( q_i \). This retrieval and refinement process is iterative, meaning that based on the analysis of \( D_i \), the system may revise the current sub-query \( q_i \) or even generate new sub-queries if the retrieved information is insufficient. Each iteration contributes to the reasoning history \( H_{i+1} \), and the sub-query decomposition continues until the reasoning path is terminated for the original query \( Q \).

In cases where retrieval is unnecessary, the system directly utilizes the internal knowledge stored in \( H_i \) to compute the answer \( \tilde{a}_i \), thereby bypassing document retrieval and speeding up the query resolution process.


The query decomposition and retrieval process can be viewed as a search over reasoning states, where each reasoning state \( s_i \) represents the current knowledge gained after processing the sub-query \( q_i \) and retrieving the corresponding documents \( D_i \). Let \( s_0 \) denote the initial state defined solely by the original query \( Q \), i.e., \( s_0 = Q \). The transition from state \( s_i \) to state \( s_{i+1} \) is defined as:
\begin{equation}
s_{i+1} = T(s_i, q_i, a_i, D_i)
\end{equation}
where \( T \) is the transition function that updates the reasoning state based on the retrieved information or internal knowledge. The system transitions through a series of reasoning states \( \{ s_0, s_1, \dots, s_n \} \), where \( s_n \) is the terminal state containing sufficient information to answer \( Q \).

The dynamic query rewriting function \( f_i \) that generates the sub-query \( q_i \) in the \( i \)-th step is defined as:
\begin{equation}
q_i = f_i(Q, H_i)
\end{equation}
This function takes into account both the original query \( Q \) and the current history \( H_i \), ensuring that the sub-query generated at each step is contextually informed by previous answers.

\subsection{Refine}

Once the review phase concludes, and all necessary information is gathered, the refine phase begins. This phase is responsible for multi-step aggregation of the intermediate answers \( \{a_1, a_2, \ldots, a_n\} \), corresponding to the sub-queries \( \{q_1, q_2, \ldots, q_n\} \). Each intermediate answer \( a_i \) contributes to the overall understanding of the query, and the system aggregates them into a coherent and unified answer.

Formally, the aggregation function \( A \) combines these intermediate answers into the final answer \( A_f \):
\begin{equation}
A_f = A(\{a_1, a_2, \ldots, a_n\})
\end{equation}
The aggregation process can be viewed as a form of information fusion, where disparate sources of information (retrieved documents, internal knowledge, reasoning history) are synthesized into a final answer.

\section{Experiments and Analysis}
\subsection{Datasets}
We experimented experiments on four datasets, which contain two major types of multi-hop question answering tasks. (1) Regular multi-hop question answering: 
the aim of this task is to sequentially reason through multiple pieces of evidence to answer a complex question. For this category, we used two established datasets: 2WikiMultiHopQA \cite{ho20202wiki} and MultiHopRAG \cite{tang2024multihoprag}.  (2) Dynamic multi-hop question answering, which includes queries where the underlying knowledge or answer is updated over time. We choose FreshQA \cite{vu2023freshllms} and PAT-Questions \cite{meem2024pat} in this task.

We report the data statistics and settings for all the datasets used in our experiments in Table~\ref{tab:data_statistics}. These datasets span both dynamic and static multi-hop question answering tasks, providing a comprehensive evaluation of the proposed method across diverse challenges. FreshQA consists of 377 questions where the correct answers may change over time as new information becomes available. 
We evaluated the dataset with 600 questions, which include a mixture of single-hop and multi-hop reasoning tasks. For PAT-Questions, 2WikiMultihopQA and MultiHopRAG, we randomly choose 500 examples for testing.

\begin{table*}[htbp]
    \centering
    \caption{Data statistics and experimental settings of the datasets used in our experiments.}
    \label{tab:data_statistics}
    \begin{adjustbox}{max width=\textwidth}
    \begin{tabular}{lcccccccc}
    \toprule
    \textbf{Settings} & \textbf{FreshQA} & \textbf{PAT-Questions} & \textbf{2WikiMultihopQA} & \textbf{MultiHopRAG} \\
    \midrule
    Task & dynamic & dynamic & regular & regular \\
    \#Examples & 600 & 6,172 & 192,606 & 2,556 \\
    Metrics & single-hop \& multi-hop Acc. & single-hop \& multi-hop Acc. & Acc., F1 & Acc., F1 \\
    Corpus & Web Sources & Wikipedia & Wikipedia & Wikipedia \\
    Top-k & 5 & 3 & 3 & 3 \\
    
    \bottomrule
    \end{tabular}
    \end{adjustbox}
\end{table*}

\subsection{Baselines}

In order to evaluate the performance of our proposed method, \textit{review-then-refine}, we compare it against several established baselines across both dynamic and static multi-hop question answering tasks. The chosen baselines cover a wide spectrum of methodologies, from conventional retrieval-based models to more recent advancements that integrate reasoning and generation techniques. Below, we briefly describe each of the baselines utilized in our experiments, as well as their relevance to dynamic and static QA settings.

\textbf{Vanilla}: This is a straightforward baseline that employs a standard pipeline approach, where questions are answered using a basic mechanism with LLMs directly. It presents tasks and the questions, and the LLM predicts the outcomes of the
questions through in-context learning.

\textbf{Freshprompt}: Freshprompt~\cite{vu2023freshllms} is designed specifically for dynamic multi-hop QA, where knowledge freshness is critical. The model leverages large language models (LLMs) with prompting techniques to retrieve and reason over recently updated knowledge. 

\textbf{Chain-of-Note}: Chain-of-Note(CoN) \cite{yu2023chainofnote} introduces summarized contextual navigation, where the model incrementally navigates through the knowledge graph or document space by conditioning each retrieval step on previously gathered evidence. 

\textbf{Self-Ask}: Self-Ask~\cite{self-askpress2022measuring} is a method where the model generates its own sub-questions, decomposing the original query into simpler queries in an autonomous manner. The model retrieves answers to each sub-question and aggregates them to form the final answer. 

\textbf{ReAct}: ReAct~\cite{yao2022react} integrates retrieval and reasoning using action-based prompts that guide the model in generating sub-questions. 

\textbf{RAT}: The RAT (Retrieval-Augmented Thoughts) \cite{wang2024rat} introduces a novel method for enhancing the reasoning and generation capabilities of large language models (LLMs) by iteratively revising their chain-of-thought (CoT)\cite{wei2022cot} with the aid of information retrieval. RAT dynamically refines the reasoning at each step, significantly improving performance on long-horizon generation tasks, where multiple reasoning steps are needed to arrive at a coherent and accurate conclusion.

\textbf{DSPy}: DSPy~\cite{khattab2024dspy} is a dynamic search planning method that uses a multi-step decision-making process to strategically retrieve relevant information and reason over it. 

\subsection{Evaluation}
Evaluating the effectiveness of multi-hop question answering models requires a comprehensive approach that assesses both the accuracy of the model's reasoning capabilities and its ability to handle dynamic and evolving knowledge. To measure the task performance of the models, we use the following widely accepted metrics: \textbf{Single-hop Accuracy}: This metric evaluates the model’s performance on individual sub-questions within the overall multi-hop task. It is especially important for dynamic tasks such as FreshQA and PAT-Questions, where the model must handle evolving information across distinct hops. \textbf{Multi-hop Accuracy}: This measures the overall accuracy for the complete reasoning task, where multiple hops are required to answer the query. This metric provides insight into how well the model integrates intermediate information to produce a final answer. \textbf{F1 Score}: For static datasets such as MultiHopRAG and 2WikiMultiHopQA, F1 is used to measure the overlap between the predicted answer and the ground-truth answer. This metric is critical for tasks where exact matches may be too strict, and partial overlaps (e.g., synonymous phrases) should be considered.

\begin{table*}[htbp]
    \centering
    \caption{Performance comparison across different datasets and methods with GPT-3.5-turbo. For dynamic tasks, we evaluate the accuracy on both single-hop and multi-hop questions, and they are shown as single-hop Acc. and multi-hop Acc. on the table respectively.}
    \label{tab:all_evaluation}
    \begin{adjustbox}{max width=\textwidth}
    \begin{tabular}{lcccccccccc}
    \toprule
    & \multicolumn{4}{c}{\textbf{Dynamic}} & \multicolumn{4}{c}{\textbf{Regular}} \\
    \cmidrule(lr){2-5} \cmidrule(lr){6-9}
    & \multicolumn{2}{c}{\textbf{FreshQA}} & \multicolumn{2}{c}{\textbf{PAT-Questions}} & \multicolumn{2}{c}{\textbf{MultiHopRAG}} & \multicolumn{2}{c}{\textbf{2Wikimultihopqa}} \\
    \cmidrule(lr){2-3} \cmidrule(lr){4-5} \cmidrule(lr){6-7} \cmidrule(lr){8-9}
    & single-hop Acc. & multi-hop Acc. & single-hop Acc. & multi-hop Acc. & Acc. & F1 & Acc. & F1 \\
    \midrule
    Vanilla & 35.47 & 28.50 & 12.10 & 7.89 & 28.57 & 30.44 & 26.85 & 34.91 \\
    Freshprompt & 68.87 & 49.52 & 23.91 & 13.58 & 42.86 & 35.71 & 37.06 & 39.28 \\
    CoN & 63.20 & 53.67 & 22.71 & 11.76 & 30.46 & 27.78 & 38.85 & 40.05 \\
    self-ask & 59.13 & 49.52 & 22.95 & 17.78 & 41.17 & 39.17 & 40.86 & 44.77 \\
    ReAct & 58.61 & 50.95 & 30.17 & 17.82 & 42.19 & 40.51 & 36.02 & 40.07 \\
    RAT & 65.39 & 40.91 & 35.29 & 26.73 & 43.06 & 38.10 & 43.66 & 43.05 \\
    DSPy & 66.57 & 50.01 & 35.84 & 25.66 & 46.44 & 46.89 & 43.51 & \textbf{45.06} \\
    \midrule
    \textit{review-then-refine} & \textbf{70.68} & \textbf{60.60} & \textbf{39.02} & \textbf{27.59} & \textbf{47.25} & \textbf{48.09} & \textbf{44.03} & 44.28 \\
    \bottomrule
    \end{tabular}
    \end{adjustbox}
\end{table*}

\begin{table*}[htbp]
    \centering
    \caption{Ablation Study Results across Dynamic and Static Multi-Hop QA Datasets.}
    \label{tab:ablation}

    \begin{adjustbox}{max width=\textwidth}
    \begin{tabular}{lcccccccccc}
    \toprule
    & \multicolumn{4}{c}{\textbf{Dynamic}} & \multicolumn{4}{c}{\textbf{Regular}} \\
    \cmidrule(lr){2-5} \cmidrule(lr){6-9}
    & \multicolumn{2}{c}{\textbf{FreshQA}} & \multicolumn{2}{c}{\textbf{PAT-Questions}} & \multicolumn{2}{c}{\textbf{MultiHopRAG}} & \multicolumn{2}{c}{\textbf{2Wikimultihopqa}} \\
    \cmidrule(lr){2-3} \cmidrule(lr){4-5} \cmidrule(lr){6-7} \cmidrule(lr){8-9}
    & single-hop Acc. & multi-hop Acc. & single-hop Acc. & multi-hop Acc. & Acc. & F1 & Acc. & F1 \\
    \midrule
    w/o query decompose & 61.05 & 48.72 & 24.98 & 15.46 & 39.85 & 37.22 & 38.23 & 37.96 \\
    w/o retrieval module & 56.14 & 43.24 & 27.59 & 16.03 & 42.41 & 48.97 & 39.86 & 38.73 \\
    w/o dynamic rewrite & 67.74 & 55.62 & 37.02 & 23.77 & 46.96 & 47.88 & 43.44 & 44.01 \\
    \bottomrule
    \end{tabular}
    \end{adjustbox}
\end{table*}

\subsection{Results}
In this section, we present the experimental results across both dynamic and static multi-hop question answering tasks and compare the performance of \textit{review-then-refine} with several baseline models. We choose \texttt{gpt-3.5-turbo} as the base model. Implementation detail is shown in Appendix~\ref{sec:implement}. The results are summarized in Table~\ref{tab:all_evaluation} and highlight key performance metrics such as single-hop accuracy, multi-hop accuracy, and F1 score. Additional results are shown in Appendix \ref{sec:addres}.

\textbf{Dynamic Tasks}: \textit{Review-then-refine} demonstrates significant improvements in dynamic multi-hop question answering tasks, which require handling evolving knowledge sources. In FreshQA, \textit{review-then-refine} achieves the highest single-hop accuracy of 70.68\% and multi-hop accuracy of 60.60\%, surpassing all baseline methods. The comparison with other strong baselines like DSPy (66.57\%) and CoN (63.20\%) shows that our method's dynamic query rewriting mechanism leads to more effective handling of multi-hop reasoning. Additionally, the multi-hop accuracy outperforms Freshprompt (49.52\%) and RAT (40.91\%) by a large margin, demonstrating the robustness of our approach in capturing and reasoning over multiple, sequentially dependent facts.

In addition, \textit{review-then-refine} achieves 39.02\% multi-hop accuracy, outperforming DSPy (35.84\%) and RAT (35.29\%) on PAT-Questions dataset. Notably, this task involves querying across multiple patent-related documents, which makes efficient retrieval and reasoning crucial. Our framework’s dynamic query refinement proves critical in this context, as it enables the system to iteratively improve its predictions as new information is gathered. In comparison, models like ReAct (30.17\%) and CoN (22.71\%) struggle to achieve comparable results, as they do not integrate retrieval as effectively with the reasoning process.

\textbf{Regular Tasks}: Our method also performs strongly on static datasets, where the knowledge sources remain fixed and the task is to reason over existing, unchanging information. On MultiHopRAG dataset, \textit{review-then-refine} achieves 47.25\% accuracy, outperforming RAT(43.06\%) and ReAct (42.19\%). The F1 score of 48.09\% further underscores our model's ability to generate precise answers, particularly in tasks where partial matches are also important. As for the results on 2WikiMultiHopQA dataset, \textit{review-then-refine} achieves 44.03\% accuracy and 44.28\% F1, making it competitive with DSPy (43.51\% accuracy and 45.06\% F1) and outperforming self-ask (40.86\% accuracy and 44.77\% F1). The performance gain can be attributed to our model’s capacity for handling complex multi-hop queries, which require extracting and synthesizing relevant information from multiple Wikipedia articles. And this results also reflects the strength of \textit{review-then-refine} in efficiently aggregating information across multiple documents without unnecessary retrieval steps.

In summary, the results show that our proposed framework, \textit{review-then-refine}, outperforms strong baselines and achieve state-of-the-art results across both dynamic and static QA datasets. And the results clearly demonstrate the superiority of \textit{review-then-refine}, particularly in dynamic and static multi-hop question answering tasks where reasoning over evolving knowledge is critical. The dynamic query rewriting mechanism allows the model to iteratively refine its reasoning, leading to more accurate and precise answers.

\subsection{Ablation Study}

To further understand the contribution of different components within \textit{review-then-refine}, we conducted an extensive ablation study across both dynamic and static multi-hop question answering tasks. The ablation results are presented in Table~\ref{tab:ablation}, where we compare the performance of the full model with several key variants. These include models without the dynamic rewrite mechanism, without the retrieval module, and without query decomposition. This study is critical for isolating the impact of individual components 
on overall task performance.

\textbf{Without Query Decomposition}
Removing the query decomposition module results in a notable drop in both single-hop and multi-hop accuracy, particularly for dynamic datasets like FreshQA and PAT-Questions. For example, the multi-hop accuracy on FreshQA decreases from 60.60\% to 48.72\%, while the multi-hop accuracy on PAT-Questions falls sharply from 27.59\% to 15.46\%. These results highlight the importance of breaking down complex queries into smaller, manageable sub-queries for efficient multi-hop reasoning. Query decomposition allows the model to focus on individual reasoning steps, which is especially vital in dynamic environments where the knowledge base is continuously updated.

\textbf{Without Retrieval Module}
Ablating the retrieval module has a significant impact on the framework’s performance, particularly for multi-hop tasks. Without the retrieval mechanism, the framework's ability to incorporate external knowledge is limited, leading to poor performance on dynamic tasks. For instance, in PAT-Questions, the multi-hop accuracy drops from 27.59\% (with retrieval) to 16.03\% (without retrieval). Similarly, on static datasets like 2WikiMultiHopQA, accuracy decreases from 44.03\% to 39.86\%. The results show that external retrieval is crucial for answering complex, multi-hop questions, particularly in cases where the information required to answer the query is not contained within the initial context.

\textbf{Without Dynamic Query Rewrite}
The dynamic query rewriting mechanism is another key component of our method. Removing this feature leads to a significant performance decrease across both dynamic and static datasets. On FreshQA, for example, multi-hop accuracy decreases from 60.60\% to 55.62\%, while in MultiHopRAG, accuracy drops from 47.25\% to 46.96\%. Dynamic query rewriting helps the model iteratively refine the retrieval and reasoning process based on intermediate answers, reducing the error accumulation that often occurs in multi-hop reasoning tasks. The slight drop in performance on static datasets further highlights the mechanism's ability to adapt and correct the reasoning path even in relatively stable environments.

The query decomposition module enables the model to break down complex questions, improving its ability to reason over multiple steps. The retrieval module is essential for integrating external knowledge, without which the model's performance degrades significantly. Finally, the dynamic query rewrite mechanism ensures the reasoning process is continually refined, leading to more accurate and efficient answers. Together, these components form a robust framework for both dynamic and static multi-hop question answering tasks, as evidenced by the superior performance of the full method.

\section{Conclusion}
In this paper, we introduced a novel framework, \textit{review-then-refine}, aimed at improving multi-hop QA with temporal information by LLMs. The framework decomposes complex queries into sub-queries, rewrites the sub-queries, retrieves relevant external information, and synthesizes it with the model’s internal knowledge, leading to more accurate, coherent and time-sensitive answers. And extensive experiments demonstrate that \textit{review-then-refine} significantly outperforms traditional \textit{retrieve-then-read} approaches, particularly in dynamic and multi-hop QA tasks. By implementing dynamic query rewriting, decomposition, and adaptive retrieval, we reduce errors and enhance the efficiency of multi-hop QA.
Ablation studies further validate the importance of each component—dynamic rewriting, retrieval, and query decomposition—in achieving state-of-the-art results across multiple datasets. In conclusion, this framework sets a new standard for handling both temporal multi-hop QA tasks and provides a foundation for future research into more advanced RAG frameworks.

\section*{Limitations}
While the \textit{review-then-refine} framework improves over traditional retrieve-then-read approaches, several limitations remain. First, the system relies on the quality of retrieved documents, and while the dynamic query rewriting and adaptive retrieval help reduce irrelevant information, external hallucinations can still occur, especially when conflicting or biased data are retrieved.

Second, the framework faces potential efficiency issues in handling very long reasoning chains. The iterative nature of query rewriting can lead to computational overhead, particularly in real-time applications where speed is crucial.

Additionally, the system’s reliance on external knowledge sources may pose challenges in fast-evolving domains. Though dynamic retrieval decisions are made, outdated internal knowledge can impact performance on time-sensitive queries.

And while the model showed strong results on datasets like FreshQA and MultiHopRAG, these do not represent the full spectrum of multi-hop QA scenarios. Specialized domains may require more refined retrieval and reasoning processes, and further optimization for real-time use remains an open challenge.

\bibliography{main}

\appendix

\section{Implementation Detail}
\label{sec:implement}
We access the GPT models through the OpenAI API, specifically utilizing the \texttt{gpt-3.5-turbo} version. The temperature for generation is set to 0.3 and  top-p to 1 to balance diversity and precision in generated outputs. All experimental evaluations use the same set of hyperparameters to ensure consistency across tasks. For evaluation metrics, we apply \textbf{Accuracy (Acc.)} for both single-hop and multi-hop questions in dynamic tasks, while in regular tasks, we measure \textbf{Accuracy (Acc.)} and \textbf{F1 score}.

Our framework integrates a retrieval mechanism to access external knowledge when required. For this, we utilize ColBERT-v2 \cite{santhanam2021colbertv2} as the retriever model to identify the most relevant documents from external corpora. The retrieval process works as follows:
\begin{itemize}
    \item \textbf{Corpus}: The retrieval process targets different corpora depending on the dataset. For regular tasks (e.g., 2WikiMultihopQA, MultiHopRAG), Wikipedia serves as the primary knowledge source. For dynamic tasks (e.g., FreshQA, PAT-Questions), we rely on a broader set of web sources, including more recently updated documents.
    \item \textbf{Retriever}: ColBERT-v2 is employed as the main retriever to rank relevant documents based on semantic similarity between the query and the document embeddings. ColBERT-v2 uses a late interaction mechanism, where the query and document are independently encoded into dense vectors. The similarity is computed through interaction between these embeddings. The retriever returns the top-k relevant documents, where \( k \) is typically set to 2 or 3 depending on the task.
\end{itemize}

During the dynamic query rewriting phase, the model selectively invokes the retriever when it deems external information necessary. The retrieval results are then integrated into the reasoning chain and synthesized with the internal knowledge of the LLMs. This approach ensures that external information complements the model’s internal knowledge, reducing the risk of hallucination and improving overall performance in multi-hop question answering tasks.

\section{Additional Results} 
\label{sec:addres} 
In this section, we evaluate the performance of various methods on dynamic and regular multi-hop question answering tasks using \texttt{GPT-4o-mini} as the base LLM. Table \ref{tab:evaluation} presents the results, highlighting single-hop and multi-hop accuracy, along with F1 scores. \textbf{Dynamic Tasks}: Our proposed \textit{review-then-refine} method demonstrates strong performance on dynamic tasks, achieving the highest multi-hop accuracy of 62.47\% on FreshQA, significantly outperforming DSPy (54.73\%) and RAT (54.24\%). In addition, it achieves 38.68\% single-hop accuracy on PAT-Questions, surpassing DSPy (36.46\%) and RAT (36.06\%). These results emphasize the efficacy of our dynamic query rewriting mechanism, which facilitates more precise multi-hop reasoning and retrieval. \textbf{Regular Tasks}: On regular tasks, \textit{review-then-refine} continues to deliver superior results, achieving 49.26\% accuracy and 48.09\% F1 score on MultiHopRAG, outperforming all baselines including DSPy (44.76\% accuracy) and ReAct (42.19\%). For 2Wikimultihopqa, our method achieves 48.01\% accuracy and 47.56\% F1 score, closely matching DSPy’s F1 score while maintaining a competitive advantage in accuracy. In summary, \textit{review-then-refine} outperforms all baseline methods across both dynamic and regular multi-hop QA tasks. The results demonstrate its effectiveness in improving reasoning and retrieval in both evolving and static knowledge environments, making it a robust solution for multi-hop question answering.

\begin{table*}[htbp]
    \centering
    \caption{Performance comparison across different datasets and methods with \texttt{GPT-4o-mini}.}
    \label{tab:evaluation}
    \begin{adjustbox}{max width=\textwidth}
    \begin{tabular}{lcccccccccc}
    \toprule
    & \multicolumn{4}{c}{\textbf{Dynamic}} & \multicolumn{4}{c}{\textbf{Regular}} \\
    \cmidrule(lr){2-5} \cmidrule(lr){6-9}
    & \multicolumn{2}{c}{\textbf{FreshQA}} & \multicolumn{2}{c}{\textbf{PAT-Questions}} & \multicolumn{2}{c}{\textbf{MultiHopRAG}} & \multicolumn{2}{c}{\textbf{2Wikimultihopqa}} \\
    \cmidrule(lr){2-3} \cmidrule(lr){4-5} \cmidrule(lr){6-7} \cmidrule(lr){8-9}
    & single-hop Acc. & multi-hop Acc. & single-hop Acc. & multi-hop Acc. & Acc. & F1 & Acc. & F1 \\
    \midrule
    Vanilla & 41.75 & 32.01 & 14.63 & 8.70 & 29.43 & 33.43 & 29.25 & 36.17 \\ 
    Freshprompt & 72.04 & 44.58 & 23.46 & 15.28 & 31.28 & 34.70 & 35.28 & 40.08 \\ 
    CoN & 60.07 & 51.15 & 21.02 & 12.50 & 32.39 & 34.42 & 40.01 & 42.14 \\ 
    Self-ask & 61.13 & 49.06 & 29.63 & 22.05 & 38.37 & 43.33 & 37.86 & 42.86 \\ 
    ReAct & 66.74 & 53.42 & 36.92 & 18.94 & 47.37 & 40.77 & 38.76 & 42.24 \\ 
    RAT & 64.55 & 54.24 & 36.06 & 25.24 & 43.11 & 40.74 & 45.86 & 44.42 \\ 
    DSPy & 66.26 & 54.73 & 36.46 & 26.23 & 44.76 & 45.68 & \textbf{48.21} & 44.07 \\ 
    \midrule
    \textit{review-then-refine} & \textbf{74.32} & \textbf{62.47} & \textbf{38.68} & \textbf{29.04} & \textbf{49.26} & \textbf{48.09} & 48.01 & \textbf{47.56} \\ 
    \bottomrule
    \end{tabular}
    \end{adjustbox}
\end{table*}

\section{Additional Experiments}
\label{sec:additional_experiments}

In this section, we present the results of additional experiments aimed at further validating the effectiveness of our proposed method under various conditions, particularly in the absence of external retrieval components. The ablation study focuses on analyzing the performance of the following baseline approaches:

\textbf{CoT \cite{wei2022cot} (Chain-of-Thought) without retrieval}: In this setting, the model relies on internally generated reasoning chains without incorporating any external knowledge through retrieval mechanisms. 

\textbf{Self-Ask \cite{self-askpress2022measuring} without retrieval}: Similar to CoT, this variant decomposes the query into sub-questions but does not utilize external retrieval to enhance the accuracy of the answers.

\textbf{SearChain \cite{xu2024search} without retrieval}: This method further breaks down the question into a series of sub-questions and answers but omits the retrieval step, relying solely on the model's internal knowledge base.

\subsection{Results}
Table~\ref{tab:dynamic_results} shows the accuracy across the dynamic multi-hop datasets, FreshQA and PAT-Questions, for both single-hop and multi-hop question types. And table~\ref{tab:static_results} shows the accuracy across the regular multi-hop datasets, MultiHopRAG and 2WikiMultihopQA, for both Accuracy (Acc.) and F1 score.

\begin{table*}[htbp]
    \centering
    \caption{Performance comparison of different ablation methods across dynamic multi-hop QA datasets.}
    \label{tab:dynamic_results}
    \begin{adjustbox}{max width=\textwidth}
    \begin{tabular}{lcccc}
    \toprule
    & \multicolumn{2}{c}{\textbf{FreshQA}} & \multicolumn{2}{c}{\textbf{PAT-Questions}} \\
    \cmidrule(lr){2-3} \cmidrule(lr){4-5}
    & single-hop Acc. & multi-hop Acc. & single-hop Acc. & multi-hop Acc. \\
    \midrule
    CoT & 40.60 & 27.74 & 22.04 & 9.50 \\
    Self-ask w.o. retrieval & 49.19 & 26.14 & 16.94 & 8.47 \\
    SearChain w.o. retrieval & 52.14 & 40.27 & 23.38 & 17.01 \\
    \textit{review-then-refine} & 56.14 & 43.24 & 27.59 & 16.03 \\
    \bottomrule
    \end{tabular}
    \end{adjustbox}
\end{table*}

\begin{table*}[htbp]
    \centering
    \caption{Performance comparison of different ablation methods across regular multi-hop QA datasets.}
    \label{tab:static_results}
    \begin{adjustbox}{max width=\textwidth}
    \begin{tabular}{lcccc}
    \toprule
    & \multicolumn{2}{c}{\textbf{MultiHopRAG}} & \multicolumn{2}{c}{\textbf{2WikiMultihopQA}} \\
    \cmidrule(lr){2-3} \cmidrule(lr){4-5}
    & Acc. & F1 & Acc. & F1 \\
    \midrule
    CoT & 36.09 & 38.06 & 31.40 & 34.04 \\
    Self-ask w.o. retrieval & 35.37 & 43.42 & 34.97 & 36.05 \\
    SearChain w.o. retrieval & 40.89 & 46.15 & 40.98 & 38.98 \\
    \textit{review-then-refine} & 42.41 & 48.97 & 41.86 & 39.73 \\
    \bottomrule
    \end{tabular}
    \end{adjustbox}
\end{table*}

\subsection{Analysis}
The results demonstrate that the CoT and Self-Ask without retrieval methods perform relatively poorly, especially on multi-hop questions, indicating that external knowledge retrieval is critical for handling complex, multi-step reasoning tasks. Notably, our framework, \textit{review-then-refine} without retrieval still outperforms these baselines, showing that even without external knowledge retrieval, the framework benefits from dynamic query rewriting and careful reasoning decomposition. However, the inclusion of retrieval, as seen in previous sections, significantly boosts performance by providing access to updated and relevant external information.

In summary, while reasoning capabilities are enhanced by internal decomposition strategies, retrieval remains essential for achieving the best performance on multi-hop, dynamic question-answering tasks. The ablation further underscores the robustness of our framework, which is designed to balance internal knowledge with external retrieval, offering flexibility depending on the task's requirements.

\section{Prompts}
\label{sec:prompts}
The sample prompts for each method are shown in this section. We describe each prompt in detail below. The sample prompts should be included with corresponding exemplars in Table~\ref{tab:data_statistics} to fit specific tasks.
\enlargethispage{2\baselineskip}
\begin{figure*}[htbp]
    \begin{tcolorbox}[colback=gray!10, colframe=black, title=Prompt 1.1: Vanilla]
    \textbf{Task:}
    
    Answer the question directly without explanation. Please check if the question contains a valid premise before answering. If the question is yes-no question, answer yes or no directly. If your internal knowledge is insufficient to answer the question, respond 'Insufficient Information'.

    \textbf{Question:}
    
    \{q\}
    
    \textbf{Answer:}
    \end{tcolorbox}
\end{figure*}

\begin{figure*}[htbp]
    \begin{tcolorbox}[colback=gray!10, colframe=black, title=Prompt 1.2: Vanilla prompt with context]
    \textbf{Task:}
    
    Below is a question followed by some context from different sources. \
    Answer the question directly without explanation. Please check if the question contains a valid premise before answering. 
    Please answer the question based on the context. The answer to the question is a word or entity. \
    If the provided information and your internal knowledge is insufficient to answer the question, respond 'Insufficient Information'.\
    If it is a yes-no question, answer yes/no directly. Answer directly without explanation.

    \textbf{Question:} 
    
    \{q\}
    
    \textbf{Context:}
    
    \{context\}
    
    \textbf{Answer:}
    \end{tcolorbox}
\end{figure*}

\begin{figure*}[htbp]
    \begin{tcolorbox}[colback=gray!10, colframe=black, title=Prompt 2: Chain-of-Thought Prompt]
    \textbf{Task:}
    Take a deep breath and let's think step by step. You will be given a question. \
    Answer the question directly step-by-step without explanation to the best of your knowledge. Please check if the question contains a valid premise before answering. 
    If your internal knowledge is insufficient to answer the question, respond 'Insufficient Information'.\
    If it is a yes-no question, answer yes/no directly. Answer directly without explanation.

    \textbf{Question:} 
    
    \{q\}
    
    \textbf{Answer:}
    
    \end{tcolorbox}
\end{figure*}

\begin{figure*}[htbp]
    \begin{tcolorbox}[colback=gray!10, colframe=black, title=Prompt 3: Freshprompt]
        \textbf{Task:}

        The question is followed by some context from different sources. Please answer the question based on the context and your internal knowledge. 
        The answer to the question is a word or entity. 
        If the provided information is insufficient to answer the question, respond 'Insufficient Information'. 
        Answer directly without explanation.

        \textbf{Question:} 

        \{q\}

        \textbf{Context:}

        \{context\}

        \textbf{Answer:}

    \end{tcolorbox}
\end{figure*}

\begin{figure*}[htbp]
    \begin{tcolorbox}[colback=gray!10, colframe=black, title=Prompt 4: Chain-of-Note]
        \textbf{Task:}
        
        1. Read the given question and Wikipedia passages to gather relevant information.\
        
        2. Write reading notes summarizing the key points from these passages with no more than 3 sentences.\
        
        3. Discuss the relevance of the given question and Wikipedia passages.\
        
        4. If some passages are relevant to the given question, directly answer the question with no more than 3 sentences to the best of the knowledge. No hallucination is allowed.\
        
        5. If no passage is relevant, answer with "No relevant passage".\
        
        6. When you are generating the final answer, starting with [Final Content].\

        \textbf{Instruction:}
        
        You should read the instructions from head to tail carefully.
        Please answer the question based on the context and your internal knowledge. 
        If the provided information is insufficient to answer the question, respond 'Insufficient Information'. 

        \textbf{Question:} 

        \{q\}

        \textbf{Passages:}

        \{passages\}

        \textbf{Answer:}

    \end{tcolorbox}
\end{figure*}

\begin{figure*}[htbp]
    \begin{tcolorbox}[colback=gray!10, colframe=black, title=Prompt 5: ReAct]
        \textbf{Role:}

        You are an intelligent assistant that uses ReAct framework to answer multi-hop questions.
        
        \textbf{Task:}
        
        Answer the following questions as best you can. You have access to the tool:

        - Search: Search for a term in Wikipedia and retrieve factual information to the input question.
        
        - Skip: Use the internal knowledge to the best and generate answer to the input question without hallucination.

        Use the following format:
        
        Question: The input question you must answer.
        
        Thought: Your reasoning process about what information is needed or what step to take next.
        
        Action: The action to take, should be one of [Search, Skip].
        
        Action Input: The specific term or query to provide to the tool.
        
        Observation: The result of the action (what the tool returns).
        ... (This Thought/Action/Action Input/Observation sequence can repeat multiple times until all necessary information is gathered)
        
        Thought: I now know the final answer after considering all the relevant information.
        
        Final Answer: The final, well-supported answer to the original question.
                
        Begin!

        \textbf{Instruction:}
        
        You should read the instructions from head to tail carefully.
        Please answer the question based on the context and your internal knowledge. 
        If the provided information is insufficient to answer the question, respond 'Insufficient Information'. 

        \textbf{Question:} 

        \{q\}

        \textbf{Answer:}

    \end{tcolorbox}
\end{figure*}

\begin{figure*}[htbp]
    \begin{tcolorbox}[colback=gray!10, colframe=black, title=Prompt 6: SearChain]
        \textbf{Role:}

        You are an intelligent assistant that uses SearChain framework to answer multi-hop questions.
        
        \textbf{Task:}
        
        Construct a global reasoning chain for this complex question.
        [Question]:"{}" and answer the question, and generate a query to the
        search engine based on what you already know at each step of the reasoning chain, starting with [Query].
        
        You should generate the answer for each [Query], starting with [Answer].
        
        You should generate the final answer for the [Question] by referring the [Query]-[Answer] pairs, starting with [Final Content].
        
        For example:

        [Question]: Who is the head coach of the team that Emre Can play for currently?
        
        [Query 1]: Which team does Emre Can play for as of March 2024?

        [Answer 1]: Emre Can currently plays for Borussia Dortmund.
        
        [Query 2]: Who is the head coach of Borussia Dortmund as of March 2024?
        
        [Answer 2]: The head coach of Borussia Dortmund as of March 2024 is Edin Terzić.
        
        [Final Content]: Emre Can currently plays for Borussia Dortmund [1], and the head coach of Borussia Dortmund as of March 2024 is Edin Terzić [2]. Therefore, the answer is Edin Terzić.

        [Question]: How many places of higher learning are in the city where the Yongle emperor greeted the person to whom the edict
        was addressed?
        
        [Query 1]: Who was the edict addressed to?
        
        [Answer 1]: the Karmapa
        
        [Query 2]: Where did the Yongle Emperor greet the Karmapa?
        
        [Answer 2]: Nanjing
        
        [Query 3]: How many places of higher learning are in Nanjing?
        
        [Answer 3]: 75
        
        [Final Content]: The edict was addressed to Karmapa [1]. Yongle Emperor greet the Karampa in Nanjing [2]. There are 75 places
        of higher learning are in Nanjing [3]. So the final answer is 75.

        \textbf{Instruction:}
        
        You should read the instructions from head to tail carefully.
        Please answer the question based on the context and your internal knowledge. 
        If the provided information is insufficient to answer the question, respond 'Insufficient Information'. 

        \textbf{Question:} 

        \{q\}

        \textbf{Answer:}

    \end{tcolorbox}
\end{figure*}

\section{Case Study}
\label{sec:case_study}
In this section, we present a series of case studies focusing on multi-hop question answering tasks. Each case involves a complex question that is decomposed into simpler, step-by-step sub-queries, allowing the model to reason through multiple pieces of information sequentially. We aim to illustrate how the system handles these multi-step queries and arrives at a final answer by combining intermediate answers from each step.

The following tables outline various case studies from different datasets (FreshQA: Table \ref{table:freshqa}, PAT-Questions: Table \ref{table:pat}, 2WikiMultihopQA: Table \ref{table:2wikimultihopqa}, MultiHopRAG: Table \ref{tab:multihoprag}), each showcasing different question types and reasoning paths. The tables highlight the original question, the sub-queries generated, and the refined answers for each step, leading to a final aggregated answer.

\newpage

\begin{table*}[htbp]
    \centering
    \caption{A Case on FreshQA}
    \label{table:freshqa}
    \begin{adjustbox}{max width=\textwidth}
    \begin{tabular}{p{\textwidth}}
    \toprule
    \textbf{Question:} How old is the most-followed user on TikTok?\\[5pt]
    
    \textbf{Decomposed sub-queries reasoning path:} \\
    \{ 
    "step 1": "Who is the most-followed user on TikTok?", 
    "step 2": "How old is this person?" 
    \}\\[5pt]
    
    \textbf{Step 1:} \\
    \textit{Review} \\
    Query Rewriting: Who is the most-followed user on TikTok as of June 2024?\\
    Answer: [need\_retrieval].\\
    \textit{Refine} \\
    Retrieve: [the most-followed user on TikTok] [June 2024] \\
    Context: From the comedic genius of Khaby Lame to actor Will Smith, here are the top 10 most-followed TikTok accounts, representing TikTok stardom and their incredible influence on the platform, as of June 2024. \\
    Top 10 most-followed TikTok influencers in 2024: \\
    Rank	Name	TikTok Official Username	Followers \\
    1	Khaby Lame	@khaby.lame	161.8M \\
    2	Charli D’Amelio	@charlidamelio	151.6M \\
    ... \\
    Refined Answer: As of June 2024, the most-followed user on TikTok is Khaby Lame. \\[5pt]

    \textbf{Step 2:} \\
    \textit{Review} \\
    Query Rewriting: How old is Khaby Lame in June 2024?\\
    Answer: Khaby Lame was born on March 9, 2000. In June 2024, Khaby Lame is 24 years old.\\
    \textit{Refine} \\
    Refined Answer: Khaby Lame was born on March 9, 2000. In June 2024, Khaby Lame is 24 years old.\\[5pt]
    
    \textbf{Question Answering:}\\
    Sub-Answer 1: As of June 2024, the most-followed user on TikTok is Khaby Lame.\\
    Sub-Answer 2: Khaby Lame was born on March 9, 2000. In June 2024, Khaby Lame is 24 years old.\\
    \textit{Aggregated Answer:} As of June 2024, the most-followed user on TikTok is Khaby Lame, he is 24 years old.\\
    \bottomrule
    \end{tabular}
    \end{adjustbox}
    \vspace{5pt}
\end{table*}

\newpage
\begin{table*}[htbp]
    \centering
    \caption{A Case on PAT-Questions}
    \label{table:pat}
    \begin{adjustbox}{max width=\textwidth}
    \begin{tabular}{p{\textwidth}}
    \toprule
    \textbf{Question:} Who is the spouse of the previous owner of Agrofert?\\[5pt]
    
    \textbf{Decomposed sub-queries reasoning path:}\\
    \{ \\
    \quad "step 1": "Who is the previous owner of Agrofert?", \\
    \quad "step 2": "Who is the spouse of the previous owner of Agrofert?" \\
    \}\\[5pt]
    
    \textbf{Step 1:}\\
    \textit{Review}\\
    Query Rewriting: Who is the previous owner of Agrofert as of March 2024?\\
    Answer: Andrej Babiš.\\
    \textit{Refine}\\
    Refined Answer: Andrej Babiš is the previous owner of Agrofert by March 2024.\\[5pt]
    
    \textbf{Step 2:}\\
    \textit{Review}\\
    Query Rewriting: Who is the spouse of Andrej Babiš in March 2024?\\
    Answer: [need\_retrieval].\\
    \textit{Refine}\\
    Retrieve: [the spouse of Andrej Babiš] [March 2024]\\
    Context: Since the 1990s Babiš has lived with his former secretary Monika Herodesová (born 1974), with whom he has two children, Vivien and Frederik. Both Monika and her brother Martin Herodes were also involved in the Stork's Nest case. In 2013 Monika changed her surname to Babišová, and they married in 2017. The couple split in April 2024.\\
    Refined Answer: As of March 2024, Andrej Babiš's spouse is Monika Babišová. However, the couple split in April 2024.\\[5pt]
    
    \textbf{Question Answering:}\\
    Sub-Answer 1: Andrej Babiš is the previous owner of Agrofert by March 2024.\\
    Sub-Answer 2: As of March 2024, Andrej Babiš's spouse is Monika Babišová. However, the couple split in April 2024.\\
    \textit{Aggregated Answer:} In March 2024, the previous owner of Agrofert is Andrej Babiš, and his spouse is Monika Babišová at that time.\\
    \bottomrule
    \end{tabular}
    \end{adjustbox}
    \vspace{5pt}
\end{table*}

\newpage
\begin{table*}[htbp]
    \centering
    \caption{A Case on 2WikiMultihopQA}
    \label{table:2wikimultihopqa}
    \begin{adjustbox}{max width=\textwidth}
    \begin{tabular}{p{\textwidth}}
    \toprule
    \textbf{Question:} Which film has the director born later, \textit{Mister Buddwing} or \textit{Road to Istanbul}?\\[5pt]

    \textbf{Decomposed sub-queries reasoning path:} \\
    \{ \\
    \quad "step 1": "Who is the director of \textit{Mister Buddwing}?", \\
    \quad "step 2": "When was the director of \textit{Mister Buddwing} born?", \\
    \quad "step 3": "Who is the director of \textit{Road to Istanbul}?", \\
    \quad "step 4": "When was the director of \textit{Road to Istanbul} born?", \\
    \quad "step 5": "Which director was born later?" \\
    \}\\[5pt]

    \textbf{Step 1:} \\
    \textit{Review} \\
    Query Rewriting: Who is the director of \textit{Mister Buddwing}?\\
    Answer: The director of \textit{Mister Buddwing} is Delbert Mann.\\
    \textit{Refine} \\
    Refined Answer: The director of \textit{Mister Buddwing} is Delbert Mann.\\[5pt]

    \textbf{Step 2:} \\
    \textit{Review} \\
    Query Rewriting: When was Delbert Mann born?\\
    Answer: Delbert Mann was born on January 30, 1920.\\
    \textit{Refine} \\
    Refined Answer: Delbert Mann was born on January 30, 1920.\\[5pt]

    \textbf{Step 3:} \\
    \textit{Review} \\
    Query Rewriting: Who is the director of \textit{Road to Istanbul}?\\
    Answer: The director of \textit{Road to Istanbul} is Rachid Bouchareb.\\
    \textit{Refine} \\
    Refined Answer: The director of \textit{Road to Istanbul} is Rachid Bouchareb.\\[5pt]

    \textbf{Step 4:} \\
    \textit{Review} \\
    Query Rewriting: When was Rachid Bouchareb born?\\
    Answer: Rachid Bouchareb was born on September 1, 1953.\\
    \textit{Refine} \\
    Refined Answer: Rachid Bouchareb was born on September 1, 1953.\\[5pt]

    \textbf{Step 5:} \\
    \textit{Review} \\
    Query Rewriting: Which director was born later, Delbert Mann or Rachid Bouchareb?\\
    Answer: Rachid Bouchareb was born later, on September 1, 1953.\\
    \textit{Refine} \\
    Refined Answer: Rachid Bouchareb was born later than Delbert Mann.\\[5pt]

    \textbf{Question Answering:}\\
    Sub-Answer 1: The director of \textit{Mister Buddwing} is Delbert Mann.\\
    Sub-Answer 2: Delbert Mann was born on January 30, 1920.\\
    Sub-Answer 3: The director of \textit{Road to Istanbul} is Rachid Bouchareb.\\
    Sub-Answer 4: Rachid Bouchareb was born on September 1, 1953.\\
    Sub-Answer 5: Rachid Bouchareb was born later than Delbert Mann.\\
    \textit{Aggregated Answer:} The director of \textit{Road to Istanbul}, Rachid Bouchareb, was born later than the director of \textit{Mister Buddwing}, Delbert Mann. Therefore, the answer is \textit{Road to Istanbul}.\\
    \bottomrule
    \end{tabular}
    \end{adjustbox}
    \vspace{5pt}
\end{table*}

\newpage
\begin{table*}[htbp]
    \centering
    \caption{A Case on MultiHopRAG}
    \label{tab:multihoprag}
    \begin{adjustbox}{max width=\textwidth}
    \begin{tabular}{p{\textwidth}}
    \toprule
    \textbf{Question:} Which company, cited in articles from both The Verge and TechCrunch, is involved in legal disputes for making exclusive deals with major tech companies, defending against accusations in an antitrust battle with a game maker, and facing a class action lawsuit for its impact on news publishers' revenues?\\[5pt]
    
    \textbf{Decomposed sub-queries reasoning path:} \\
    \quad "step 1": "Which company is involved in legal disputes for making exclusive deals with major tech companies?", \\
    \quad "step 2": "Which company is defending against accusations in an antitrust battle with a game maker?", \\
    \quad "step 3": "Which company is facing a class action lawsuit for its impact on news publishers' revenues?", \\
    \quad "step 4": "Which company is cited in articles from both The Verge and TechCrunch related to these legal disputes?" \\[5pt]

    \textbf{Step 1:} \\
    \textit{Review} \\
    Query Rewriting: Which company is involved in legal disputes for making exclusive deals with major tech companies?\\
    Answer: Google.\\
    \textit{Refine} \\
    Refined Answer: Google is involved in legal disputes for making exclusive deals with major tech companies.\\[5pt]

    \textbf{Step 2:} \\
    \textit{Review} \\
    Query Rewriting: Which company is defending against accusations in an antitrust battle with a game maker?\\
    Answer: Google is defending against Epic Games in an antitrust battle.\\
    \textit{Refine} \\
    Refined Answer: Google is defending against accusations in an antitrust battle with Epic Games.\\[5pt]

    \textbf{Step 3:} \\
    \textit{Review} \\
    Query Rewriting: Which company is facing a class action lawsuit for its impact on news publishers' revenues?\\
    Answer: Google.\\
    \textit{Refine} \\
    Refined Answer: Google is facing a class action lawsuit for its impact on news publishers' revenues.\\[5pt]

    \textbf{Step 4:} \\
    \textit{Review} \\
    Query Rewriting: Which company is cited in articles from both The Verge and TechCrunch related to these legal disputes?\\
    Answer: Google.\\
    \textit{Refine} \\
    Refined Answer: Google is cited in articles from both The Verge and TechCrunch related to these legal disputes.\\[5pt]

    \textbf{Question Answering:}\\
    Sub-Answer 1: Google is involved in legal disputes for making exclusive deals with major tech companies.\\
    Sub-Answer 2: Google is defending against accusations in an antitrust battle with Epic Games.\\
    Sub-Answer 3: Google is facing a class action lawsuit for its impact on news publishers' revenues.\\
    Sub-Answer 4: Google is cited in articles from both The Verge and TechCrunch related to these legal disputes.\\
    \textit{Aggregated Answer:} Considering all the information, the company involved in legal disputes over exclusive deals with major tech companies, facing an antitrust battle with a game maker, and facing a class action lawsuit for its impact on news publishers' revenues, cited in both The Verge and TechCrunch, is Google.\\
    \bottomrule
    \end{tabular}
    \end{adjustbox}
    \vspace{5pt}
\end{table*}

\end{document}